\documentclass[10pt,twocolumn,letterpaper]{article}

\usepackage{cvpr}
\usepackage{times}
\usepackage{epsfig}
\usepackage{graphicx}
\usepackage{amsmath}
\usepackage{amssymb}

\usepackage{todo}
\usepackage{color}

\usepackage[caption=false,font=scriptsize]{subfig}
\usepackage[linesnumbered,lined,boxed,commentsnumbered]{algorithm2e}

\DeclareMathOperator*{\argmin}{arg\,min}

\cvprfinalcopy 

\def\cvprPaperID{7} 

\SetAlCapSkip{0.2cm}
\begin{document}

\title{Face Detection, Bounding Box Aggregation and Pose Estimation for Robust Facial Landmark Localisation in the Wild}

\author{Zhen-Hua Feng$^{1,2}$~~Josef Kittler$^1$~~Muhammad Awais$^1$~~Patrik Huber$^1$~~Xiao-Jun Wu$^2$\\
$^1$ Centre for Vision, Speech and Signal Processing, University of Surrey, Guildford GU2 7XH, UK\\
$^2$School of Internet of Things Engineering, Jiangnan University, Wuxi 214122, China\\
{\tt\small \{z.feng, j.kittler, m.a.rana, p.huber\}@surrey.ac.uk, wu\_xiaojun@jiangnan.edu.cn}
}

\maketitle

\begin{abstract}
We present a framework for robust face detection and landmark localisation of faces in the wild, which has been evaluated as part of `the 2nd Facial Landmark Localisation Competition'. The framework has four stages: face detection, bounding box aggregation, pose estimation and landmark localisation. To achieve a high detection rate, we use two publicly available CNN-based face detectors and two proprietary detectors. We aggregate the detected face bounding boxes of each input image to reduce false positives and improve face detection accuracy. A cascaded shape regressor, trained using faces with a variety of pose variations, is then employed for pose estimation and image pre-processing. Last, we train the final cascaded shape regressor for fine-grained landmark localisation, using a large number of training samples with limited pose variations. The experimental results obtained on the 300W and Menpo benchmarks demonstrate the superiority of our framework over state-of-the-art methods.

\end{abstract}

\section{Introduction}
Facial landmarks provide important information for face image analysis such as face recognition~\cite{beveridge2015report,song2016dictionary,Sun_2014_CVPR,Taigman2014}, expression analysis~\cite{eleftheriadis_accv,eleftheriadis2015discriminative,happy2015automatic} and 3D face reconstruction~\cite{hu2017efficient,huber2015fitting,kittler20163d,Piotraschke_2016_CVPR,zhu2015discriminative}.
Given an input face image, the task of facial landmark localisation is to obtain the coordinates of a set of pre-defined facial landmarks.
These facial landmarks usually have specific semantic meanings, such as the nose tip and eye corners, and are instrumental in enabling the subsequent face image analysis.

The most well-known facial landmark localisation algorithms, which emerged from the early research efforts on this topic, are based on Active Shape Model (ASM)~\cite{Cootes1995}, Active Appearance Model (AAM)~\cite{cootes1998active} and Constrained Local Model (CLM)~\cite{Cristinacce2006}.
These algorithms perform well in constrained scenarios, but cannot cope reliably with faces obtained in unconstrained conditions. The current research attempts to redress this situation. Its aim is to design algorithms for robust and accurate face landmarking of unconstrained faces in the presence of extreme appearance variations in pose, expression, illumination, makeup, occlusion etc.
To this end, the recent trend has been to resort to discriminative approaches such as Cascaded Shape Regression (CSR)~\cite{Cao2014face, dollar2010cascaded, feng2015cascaded, FENG2015, daccsr_feng_2017, xiong2013supervised} and Deep Neural Networks (DNN)~\cite{Sun2013, Trigeorgis_2016_CVPR, Xiao2016, Zhang_2016_CVPR, zhang2016joint}.
Both have shown promising landmarking results for faces in the wild.

To evaluate the performance of facial landmark localisation algorithms, a number of face datasets have been collected or annotated, such as the IMM~\cite{IMM2004-03160}, BioID~\cite{jesorsky2001robust}, XM2VTS~\cite{Messer1999}, LFPW~\cite{Belhumeur2011}, COFW~\cite{Burgos-Artizzu2013}, AFLW~\cite{aflw_dataset_2011}, HELEN~\cite{le2012interactive} and 300W~\cite{sagonas2016300} datasets.
However, their key characteristics such as dataset size, appearance variation types, number of landmarks and landmark quality are not always all satisfactory or well defined.
To comprehensively evaluate and compare facial landmark localisation algorithms, the Menpo benchmark has been conceived~\cite{menpo2017cvprw,Trigeorgis_2016_CVPR}. It contains 8979 training images and 7281 test images, which have been semi-automatically annotated with 68 facial landmarks for semi-frontal faces and 39 landmarks for profile faces.
The Menpo images were selected from the AFLW~\cite{aflw_dataset_2011} and FDDB~\cite{jain2010fddb} datasets, exhibiting a wide range of appearance variations.

In contrast to the previous evaluation campaigns, the Menpo benchmark does not provide face bounding boxes for landmark initialisation, which better reflects practical applications.
\begin{figure*}[t]
\centering
	\includegraphics[trim = 0mm 85mm 35mm 0mm, clip,width=1\linewidth]{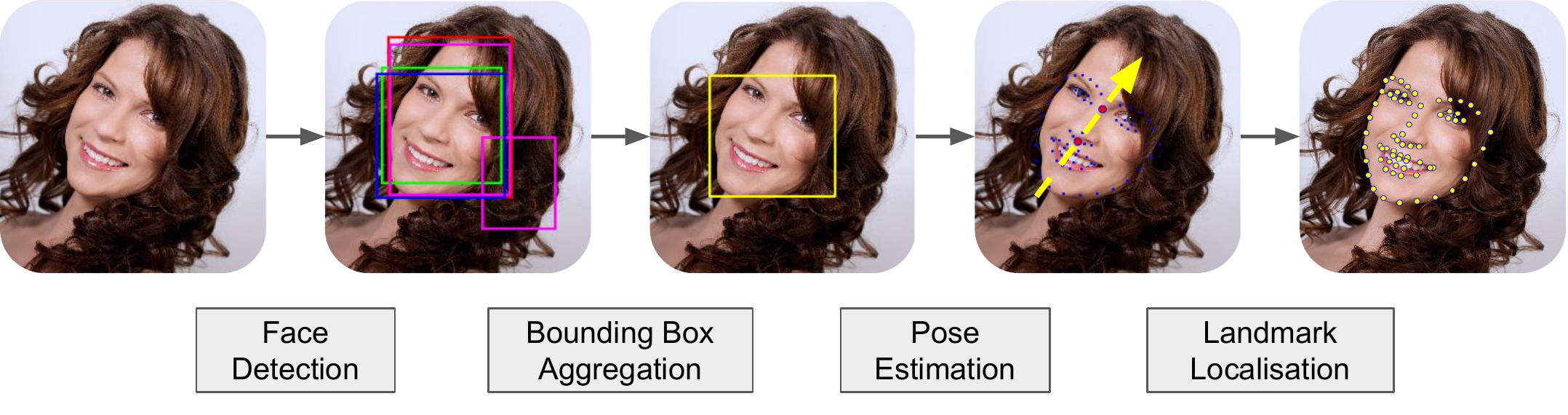}
\caption{The pipeline of our proposed facial landmark localisation framework.}
\label{fig_1}
\end{figure*}
However, the accuracy of landmark localisation algorithms highly relies on the consistency of the detected face bounding boxes for every input image. To achieve robust face detection and landmark localisation, we develop a coarse-to-fine framework designed to enhance the consistency of detected face designation as well as the accuracy of facial landmarking.
The proposed framework has four stages including face detection using multiple face detectors, face bounding box aggregation, pose estimation and facial landmark localisation, as shown in Fig.~\ref{fig_1}.
The key innovation of the proposed framework is twofold: 
1) we propose the use of multiple face detection experts to provide input to a bounding box aggregation strategy to improve the accuracy of face detection; 
2) we divide the original CSR method into a number of coarse-to-fine steps that further improve the accuracy of facial landmark localisation.

The paper is organised as follows. 
We first overview the related literature and introduce the basic cascaded shape regression method in Sections~\ref{sec_2} and \ref{sec_3}.
Then the proposed coarse-to-fine framework and its four main stages are presented in Section~\ref{sec_4}.
In Section~\ref{sec_8}, we report the experimental results obtained on the 300W and Menpo benchmarks.
Conclusions are drawn in Section~\ref{sec_9}.

\section{Related Work}
\label{sec_2}
The early advances in facial landmark localisation were mainly driven by ASM, AAM and CLM, in which a generative PCA-based shape model is used for face shape representation.
Typically in these methods, the statistical distribution of the samples used for the shape model training provides a prior for the model fitting process.
To fit the shape model to an input image, an objective function is designed to optimise the model parameters.
For example, AAM uses the intensity difference between an input image and a texture model as the loss for model optimisation.
These models and their extensions have achieved great success in facial landmarking for faces with controlled appearance variations~\cite{alabort2016unified,cootes2012robust,taam_feng_2016,feng2012automatic,baam_GonzalezMora_2007_iccv,aam_matthews_2004_ijcv,Sauer2011}.
However, for faces in the wild, the PCA-based shape model may miss some shape details and in consequence it is not able to represent complex shapes faithfully.
To address this issue, cutting-edge techniques use learning-based discriminative methods that employ the non-parametric Point Distribution Model (PDM)~\cite{pdm_1995_cootes} for shape representation.
Given an input face image and an initial shape estimate, a discriminative model learns a mapping function from the shape-related features to a shape update, driving the landmarks of the initial shape estimate to their optimal locations.

In recent years, discriminative models, in particular CSR-based methods, have become the state-of-the-art in robust facial landmark localisation of unconstrained faces.
The key to the success of CSR is the architecture cascading multiple weak regressors, which greatly improves the generalisation capacity and accuracy of a single regression model.
To form a CSR, linear regression~\cite{FENG2015, xiong2013supervised, Yan2013}, random forests or ferns~\cite{Cao2014face, ren2014face} and deep neural networks~\cite{Sun2013, Trigeorgis_2016_CVPR, Zhou2013CNN} have been used as weak regressors.
To further improve landmark localisation accuracy of CSR-based approaches, new architectures have been proposed.
For example, Feng \etal proposed to fuse multiple CSRs to improve the accuracy of a single CSR model~\cite{FENG2015}.
Xiong \etal proposed the global supervised descent method using multiple view-based CSRs to deal with the difficulties posed by extreme appearance variations~\cite{gdm_2015_xiong}.
Moreover, data augmentation~\cite{feng2015cascaded, daccsr_feng_2017, zhu2016face} and 3D-based approaches~\cite{zhu2016face}  have also been used to enhance existing CSR-based facial landmark localisation approaches.

\section{Cascaded Shape Regression}
\label{sec_3}
Given a face image $\mathbf{I}$, the face shape is represented in the form of a vector consisting of the coordinates of $L$ pre-defined 2D facial landmarks, \ie $\mathbf{s} = [x_1, ..., x_L, y_1, ..., y_L]^T$.
To automatically obtain the face shape for the input image, CSR is used in our proposed framework.
In fact, CSR is a strong regressor, $\Phi$, formed by a set of weak regressors:
\begin{equation}
\Phi = \{\phi_{1}, ..., \phi_{M}\},
\end{equation}
where $\phi_{m}$ is the $m$th weak regressor and $M$ is the number of weak regressors.
It should be noted that any regression method can be used as a weak regressor in CSR.
In this paper, we adopted ridge regression for each weak regressor.

Suppose we have $N$ training samples, $\{\mathbf{I}_n, \mathbf{s}^*_n\}_{n=1}^{N}$, where $\mathbf{I}_n$ is the $n$th training image and $\mathbf{s}^*_n$ is the ground truth face shape. We construct a CSR by progressively training a set of cascaded weak regressors.
To this end, we first initialise the face shape estimate, $\mathbf{s}'_n$, for each training image by scaling and translating the mean shape so that it perfectly fits into the detected face bounding box.
Next, the shape difference between the current shape estimate and the ground truth shape of each training sample is calculated, \ie $\delta \mathbf{s}^*_n = \mathbf{s}^*_n - \mathbf{s}'_n$.
Then the first weak regressor, $\phi_1 = \{\mathbf{A}_1, \mathbf{e}_1\}$, is obtained by solving the optimisation problem:
\begin{equation}
\argmin_{\mathbf{A}_1, \mathbf{e}_1} \sum_{n = 1}^{N}
||\mathbf{A}_1 \cdot f(\mathbf{I}_n, \mathbf{s}'_n) + \mathbf{e}_1 - \delta\mathbf{s}^*_n ||^2_2 + \lambda ||\mathbf{A}_1||_F^2,
\label{eq:csr-loss}
\end{equation}
where $\mathbf{A}_1 \in \mathbb{R}^{2L \times N_f}$ is the trained projection matrix for the first weak regressor, $N_f$ is the dimensionality of the extracted shape-related features using $f()$ and $\mathbf{e}_1 \in \mathbb{R}^{2L}$ is the offset.
$f(\mathbf{I}, \mathbf{s}')$ is a mapping function that extracts local features, \eg HOG, LBP or SIFT, from the neighbourhood around each landmark of the current shape estimate, $\mathbf{s}'$, and concatenates them into a long vector.
Solving Eq.~(\ref{eq:csr-loss}) is a typical least square estimation problem that has a closed-form solution.
Once we obtain the first weak regressor, we use it to predict the shape update,
\begin{equation}
\label{equ_s_update}
\delta \mathbf{s} = \mathbf{A} \cdot f(\mathbf{I},\mathbf{s}') + \mathbf{e},
\end{equation}
for each training sample.
Last, the current shape estimates of all the training samples are updated for the second weak regressor training, \ie $\mathbf{s}'_n \leftarrow \mathbf{s}'_n + \delta \mathbf{s}_n$ $(n = 1, ..., N)$.
We repeat this process several times until all the $M$ weak regressors are learnt.

For a test image, we first initialise the current shape estimate using the detected face bounding box as mentioned above.
Then the pre-trained weak regressors are progressively applied to the current shape estimate for shape update, as summarised in Algorithm~\ref{alg_1}.
\begin{algorithm}[t]
 \label{alg_1}
\caption{Cascaded shape regression for facial landmark localisation.}
\SetKwInOut{Input}{input}
\SetKwInOut{Output}{output}
 \Input{image $\mathbf{I}$ and a trained CSR model $\Phi = \{ \phi_1, ..., \phi_M \}$}
 \Output{facial landmarks $\mathbf{s}'$}
 initialise the current face shape, $\mathbf{s}'$, using the detected face bounding box\;
 \For{$m\leftarrow 1$ \KwTo $M$}{
 extract shape-related features, $f(\mathbf{I}, \mathbf{s}')$, from the image using the current shape estimate\;
 apply the $m$th weak regressor to obtain the shape update $\delta \mathbf{s}$ using Eq.~(\ref{equ_s_update})\;
 update the current shape estimate $\mathbf{s}' \leftarrow \mathbf{s}' + \delta \mathbf{s}$\;
 }
\end{algorithm}

\section{The Proposed Framework}
\label{sec_4}
The proposed facial landmark localisation framework has four main stages, as shown in Fig.~\ref{fig_1}.
Because the Menpo benchmark does not provide face bounding boxes for landmark initialisation, we first perform face detection for a given image.
To improve the detection rate, we use four different face detectors.
Second, the detected face bounding boxes are filtered and fused using an aggregation algorithm to reduce false positives and improve the accuracy.
Third, we predict the pose of the face image for pre-processing, \eg image rotation and flipping.
Last, cascaded shape regression is used to obtain the final facial landmark localisation result.

\begin{figure*}[t]
\centering
 \includegraphics[trim = 0mm 55mm 0mm 0mm, clip,width=1\linewidth]{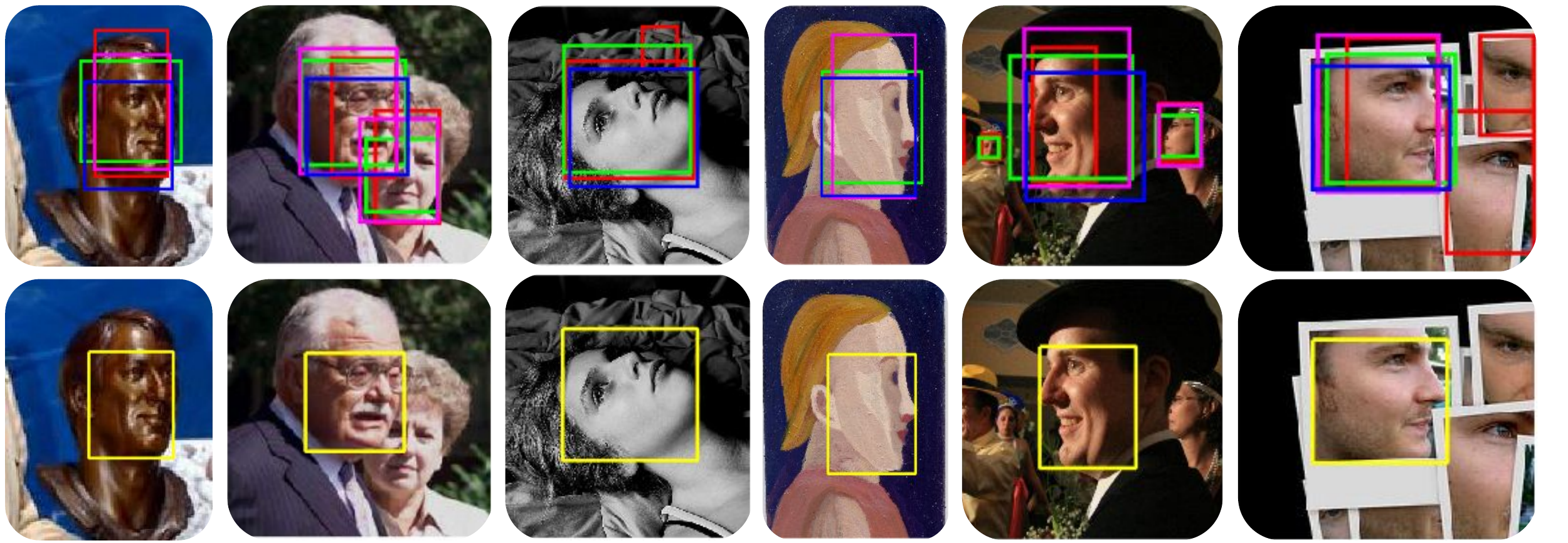}
\caption{Face bounding box aggregation: the first row shows the initial face detection results using the dlib (green), MTCNN (magenta), our own trained Faster R-CNN (red) and regression-based (blue) face detectors; the second row shows the final aggregated face bounding box for each image.}
\label{fig_2}
\end{figure*}

\subsection{Face Detection}
\label{sec_4_1}
As aforementioned, our face detection stage uses four face detectors.
The first two are public available face detectors: the Convolutional Neural Network (CNN-) based dlib face detector~\footnote{http://dlib.net} and the Multi-Task CNN (MTCNN) face detector~\cite{zhang2016joint}.
In addition, we train two proprietary face detectors to further improve the detection rate.
One of them is a general deep-learning-based face detector and the other one is a regression-based face detector specifically tailored for the Menpo benchmark.

Recently, the state of the art in deep-learning-based object detection advanced significantly~\cite{sermanet2013overfeat,he2014spatial, ren2015faster}. 
Region-based methods like Faster Region-based CNN (Faster R-CNN)~\cite{ren2015faster} and its variants emerged as state-of-the-art in detection.
Besides the two publicly available CNN-based face detectors, we train a new deep-learning-based face detector using Faster R-CNN.
Faster R-CNN consists of two main components: Region Proposal Network (RPN) and Fast R-CNN~\cite{girshick2015fast}.
RPN takes an image as input and generates rectangular object proposals, each with probability of belonging to foreground objects (objectiveness) vs background.
The RPN proposals are passed through non-maximum suppression and then sorted out with respect to objectiveness. 
Then top-N rectangular proposals are fed into Fast R-CNN for classification into different categories and refinement of the bounding box for these proposals.
For face detection, Fast R-CNN is trained to detect only one class, \ie human faces. 
We use VGG 16 layers network~\cite{simonyan2014very} as shared convolutional feature maps and use approximate joint training to update parameters.

We train our face detector using the WIDER FACE dataset~\cite{yang2016wider}. 
In order to adapt to the characteristics of the  challenge, we remove all the tiny faces (faces consisting of less that 80 pixels area) from the training set of WIDER FACE.
Furthermore, the trained face detector is fine-tuned on the training set of the Menpo benchmark.
In the challenge, a face covers a significant area of the image, therefore, we use regions with intersection-over-union (IoU) of more than 0.75 with the ground truth box as positive samples, and regions with IoU of less than 0.35 with the ground truth as negative samples in the fine tuning stage. 
Our Faster-R-CNN-based face detector has achieved very competitive results on the well-known Face Detection Dataset and Benchmark (FDDB)~\cite{jain2010fddb}.

When testing on the training set of the Menpo benchmark, even using these three deep-learning-based face detectors, it is not always possible to detect all the faces in an input image.
To address this issue, we develop a 4th face detector optimised for the Menpo benchmark by leveraging the benchmark rules, namely that `each image contains a single annotated face and it is the one that is closer to the centre of the image'.
This face detector is regression-based, which is similar to the face bounding box refinement step used in the Dynamic Attention-Controlled CSR (DAC-CSR)~\cite{daccsr_feng_2017}.
The difference is that DAC-CSR uses only one regression model for bounding box refinement, but we cascade multiple weak regressors for face bounding box regression.

The training of this cascaded bounding box regression, 
\begin{equation}
\Phi_b = \{\phi_{b,1}, ..., \phi_{b,M}\},
\end{equation}
is similar to the training of a CSR introduced in Section~\ref{sec_3}.
Here, each weak regressor updates the current face bounding box estimate, $\mathbf{b} = [x,y,w,h]'$, instead of the current face shape, $\mathbf{s}$, for an image, where $(x,y)$, $w$ and $h$ are the coordinates of the left-upper corner, width and hight of a face bounding box.
For face bounding box initialisation, we simply use the bounding box covering the whole region of an input image.
The ground truth bounding box of a face is obtained by using the one that exactly encloses all the landmarks of the face.
As features $f(\mathbf{I}, \mathbf{b})$, we extract multi-scale HOG and LBP descriptors from the image patch inside the current face bounding box.
This regression-based face detector is trained on the training set of the Menpo benchmark using 2 weak regressors.

\subsection{Face Bounding Box Aggregation}
\label{sec_4_2}
The use of multiple face detectors results in a number of face bounding boxes as output, including false positives.
However, the Menpo benchmark protocol specifies the face closer to the centre of an input image as the correct one.
To improve the accuracy of the face detection module and reduce false positives, we perform face bounding box aggregation of all the detected faces of an input image.
Fig.~\ref{fig_2} shows some examples of our initially detected faces and the final bounding boxes based on our aggregation approach.

To perform face bounding box aggregation, we first filter the face bounding boxes detected by the first three deep-learning-based face detectors.
In this step, we eliminate all the detected face bounding boxes that satisfy one of the following conditions:
1) the detection confidence score is lower than 0.85; 2) the bounding box height is smaller than $1/5$ of the image height; and 3) the bounding box does not include the image centre.
After that, we perform face bounding box refinement for the remaining face bounding boxes output by the first three face detectors.
The face bounding box refinement step is the same as the one used in DAC-CSR~\cite{daccsr_feng_2017}.
However, we train three of them for the three deep-learning-based face detectors separately.
The training is similar to that of our 4th regression-based face detector, as introduced in the last section.
However, here we use the detected face bounding box of a face detector, rather than the one covering the whole image, for bounding box initialisation.
The refined bounding boxes are averaged to obtain the final result.
If all the detected face bounding boxes are eliminated during the filtering stage, we use the one with the highest score among all the original bounding boxes output by the dlib and MTCNN face detectors for bounding box refinement.
Last, if dlib and MTCNN have not detected any faces, we just simply use the bounding box output from our 4th regression-based face detector as the final face detection result.
\subsection{Pose Estimation}
\label{sec_4_3}
The Menpo benchmark has two test sets, semi-frontal and profile, that are evaluated separately.
These two test sets consist of face images with a variety of in-the-plane pose rotations.
In addition, the profile subset has either left- or right-rotated face images with respect to yaw angle rotation.
Based on the face bounding boxes output from the first two stages of our framework, we could train a model using a dataset with a wide range of in-the-plane and out-of-plane pose variations for facial landmark localisation.
The resulting model would cope well with pose variations due to the property of the training data.
However, the accuracy of facial landmark localisation using a model trained on such an extreme range of face poses is limited.
To mitigate this problem, we estimate the face pose in each input image for image pre-processing before the last facial landmark localisation stage.
\begin{figure}[t]
\centering
	\includegraphics[trim = 0mm 65mm 168mm 0mm, clip,width=1\linewidth]{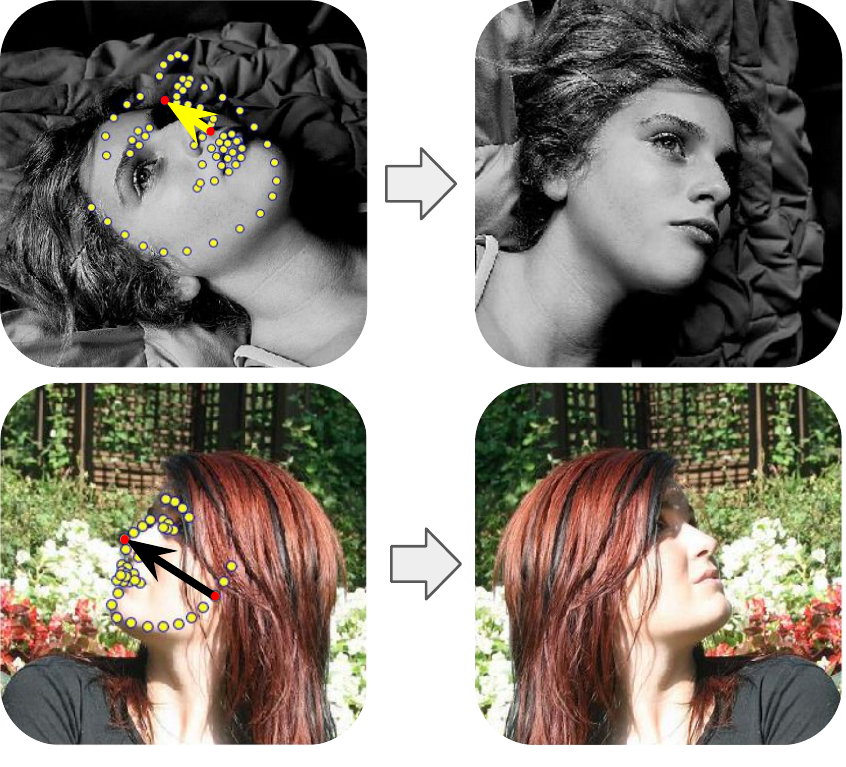}
\caption{Pose estimation and image pre-processing for semi-frontal (first row) and profile (second row) faces.}
\label{fig_3}
\end{figure}

In the pose estimation stage, we first perform rough facial landmark localisation by cascading only 2 weak regressors.
Then, for a semi-frontal face image, we use two landmarks (ID-28 and ID-34) to determine its in-the-plane rotation.
For a profile face, we use the landmarks with ID-3 and ID-20 to determine the yaw rotation.
Once the pose is estimated, we use this information to apply an appropriate pose normalisation.
We rotate the image when the estimated pose is larger than 45$^{\circ}$ away from upright for semi-frontal faces, and flip right-rotated profile face images from right to left to obtain left-rotated faces.
Two examples are shown in Fig.~\ref{fig_3}.

Because a semi-frontal face has 68 landmarks and a profile face has only 39 landmarks, we train a separate CSR-based facial landmark localiser for each of the scenarios.
To prepare the training data for the semi-frontal subset, we randomly rotate all the training images between $[0^{\circ}, 360^{\circ}]$.
For the profile subset, we flip each training image from left to right and randomly rotate it between $[-30^{\circ}, 30^{\circ}]$.
For feature extraction, we adopt both HOG and LBP descriptors computed in the neighbourhoods of all the landmarks of the current shape estimate and concatenate them into a long vector.
In addition, we also extract dense local features from the image patch enclosing all the landmarks as the contextual information, which has also been used in DAC-CSR~\cite{daccsr_feng_2017}.
\subsection{Facial Landmark Localisation}
\label{sec_4_4}
In the last stage, we perform facial landmark localisation for the face detected in an input image, using a pre-trained CSR model.
We train separate CSR models for the semi-frontal and profile subsets, using datasets with limited pose variations.
For the semi-frontal subset, we flip an original training image and perform random image rotation between $[-30^{\circ}, 30^{\circ}]$.
For the profile training faces, we flip all the right-rotated faces in yaw from right to left to construct a training set with only left-rotated profile faces.
We also perform in-the-plane rotation to all the profile faces for data augmentation.
The training and testing phases of these two CSR models are described in Section~\ref{sec_3}.
To further improve the robustness of the final localised facial landmarks, we perform random perturbation to a detected face bounding box by randomly translating the left-upper and right-bottom corners inside 5\% of the width and height of the bounding box.
We apply a pre-trained CSR to each randomly perturbed face bounding box to obtain multiple facial landmark localisation results and use the mean as the final result.
In this stage, we cascade 6 weak regressors in a CSR.
For shape-related feature extraction, \ie $f(\mathbf{I},\mathbf{s})$, we use the same technique as described in Section~\ref{sec_4_3}.

It should be noted that, for images that have been rotated or flipped in the pose estimation stage, the localised facial landmarks have to be transformed back to their original coordinate systems.
\section{Experimental Results}
\label{sec_8}
\subsection{Implementation Details}
To evaluate the performance of our proposed framework, the 300W and Menpo benchmarks were used~\cite{menpo2017cvprw}\footnote{https://ibug.doc.ic.ac.uk/resources}.
The latter one was used in the 2nd facial landmark localisation competition.
For training, the 300W benchmark provides a number of annotated face datasets including XM2VTS~\cite{Messer1999}, LFPW~\cite{Belhumeur2011}, HELEN~\cite{le2012interactive} and AFW~\cite{zhu2012face}.
For evaluation, 300W provides 300 indoor and 300 outdoor face images.
Each 300W face image has 68 facial landmarks that were semi-automatically annotated.
The Menpo benchmark has 8979 training images (6679 semi-frontal and 2300 profile faces) and 7281 test images (5335 semi-frontal and 1946 profile faces).
For each Menpo semi-frontal/profile face, 68/39 landmarks were annotated.

The 4th face detector and the face bounding box refinement models used in the aggregation stage were trained using all the training images of the Menpo benchmark.
We used the LFPW, AFW and HELEN datasets and the training set of the Menpo benchmark for the CSR training in our pose estimation and facial landmark localisation stages.
To extract HOG and LBP features, the VLFeat toolbox was used~\cite{vedaldi08vlfeat}.

\begin{figure}[!t]
\centering
\subfloat[68 points]{
\label{fig_8_1}
 \includegraphics[width=1\linewidth]{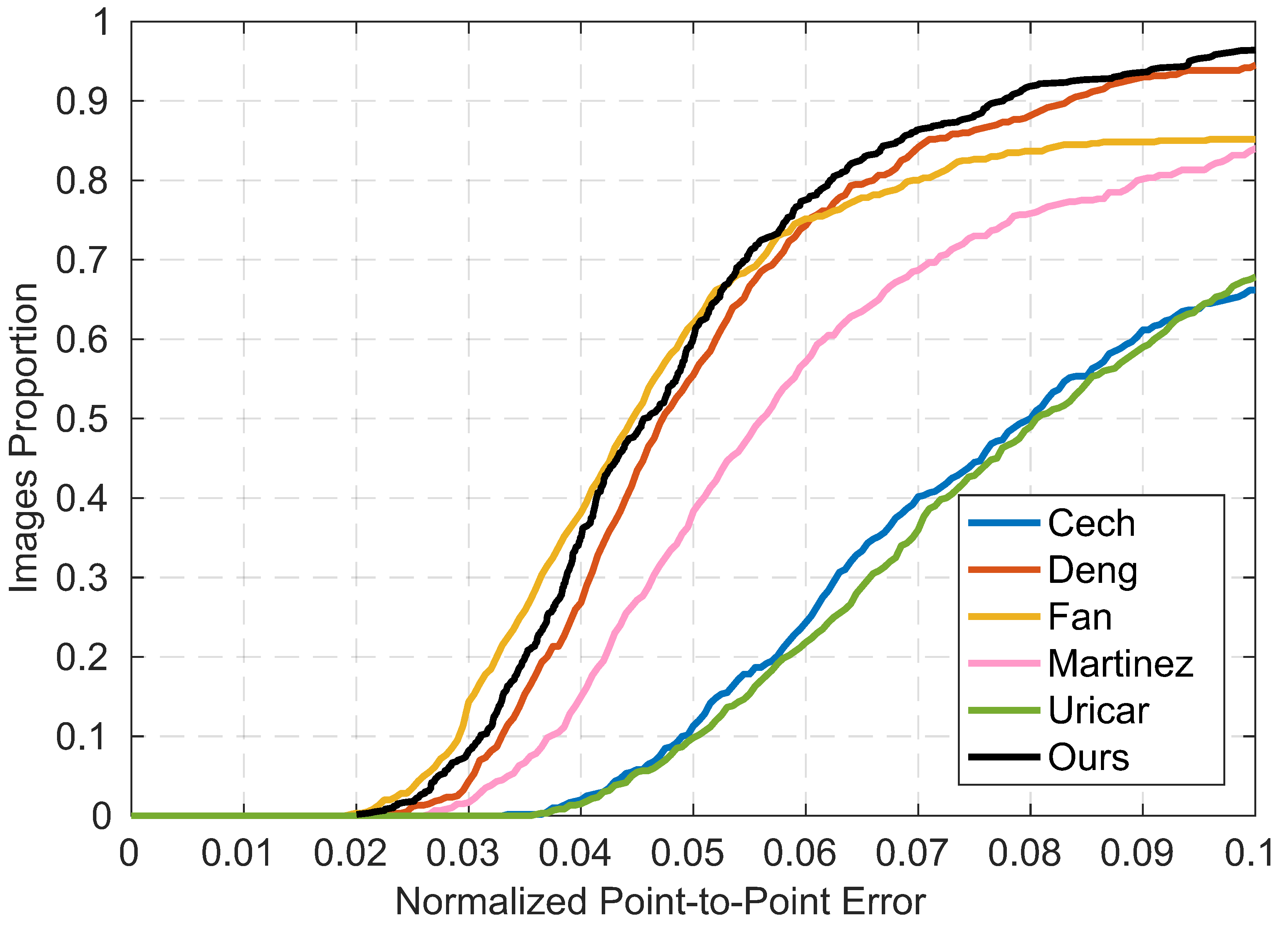}
}
\\
\subfloat[51 points]{
\label{fig_8_2}
 \includegraphics[width=1\linewidth]{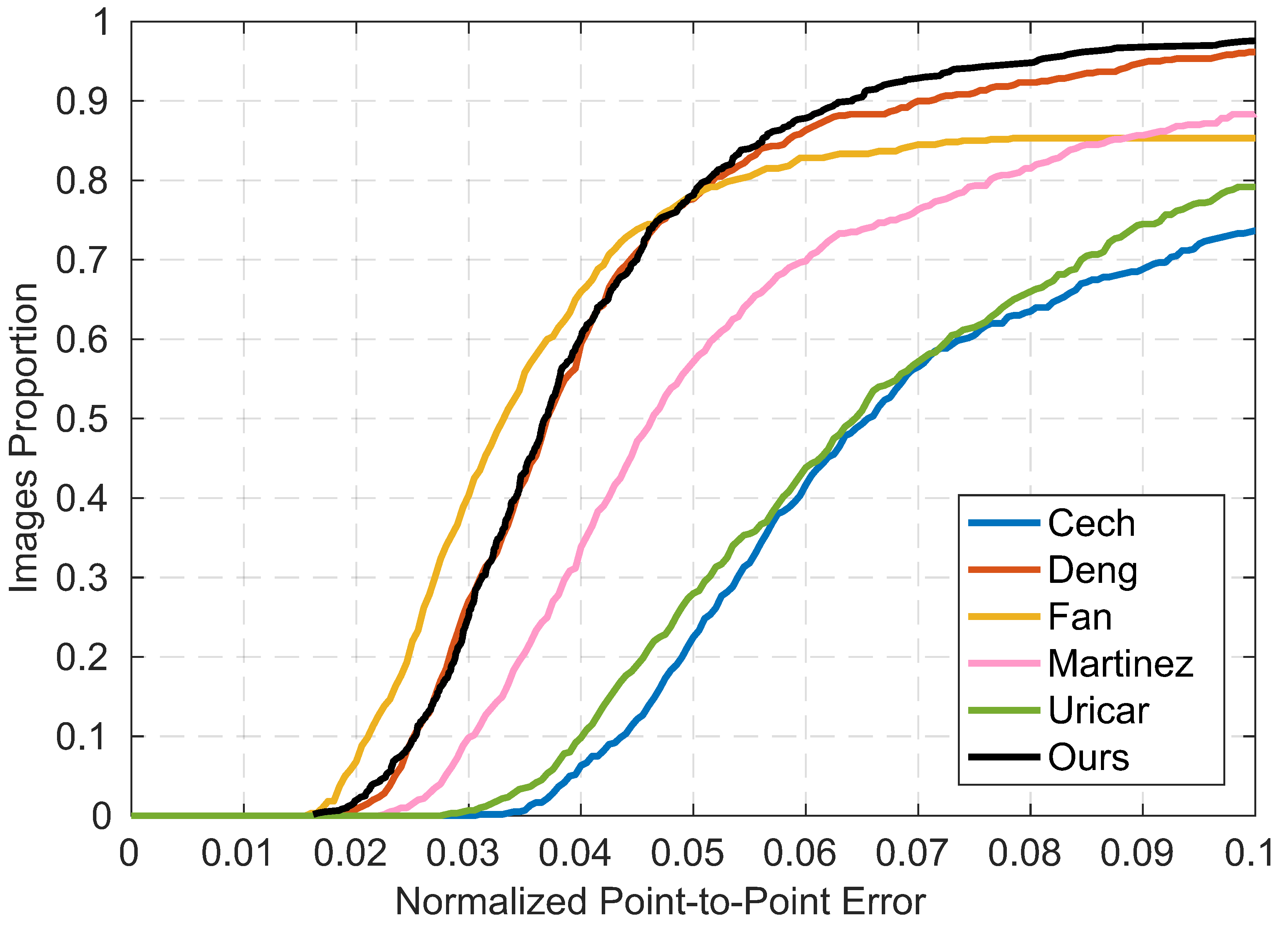}
}
\caption{A comparison of our proposed framework with state-of-the-art methods on the 300W dataset (indoor + outdoor).
The performance was evaluated in terms of the normalised RMS error over (a) 68 and (b) 51 facial points.}
\label{fig_8}
\end{figure}
\subsection{Evaluation on 300W}
We first evaluate the accuracy of our proposed framework on the 300W dataset using the protocol of the second run of the 300W challenge that was completed at the beginning of 2015~\cite{sagonas2016300}.
In general, the protocol of a facial landmark localisation benchmarking dataset provides face bounding boxes for initialisation.
The re-run of the 300W challenge is the only one that has the same protocol as the Menpo benchmark, \ie a participant has to perform face detection and landmark localisation jointly.

We compare our method with all the participants' approaches from the second evaluation campaign of the 300W challenge in terms of accuracy~\cite{sagonas2016300}.
The evaluation results are shown in Fig.~\ref{fig_8}.
The proposed framework outperforms the algorithms of all the participants in the second evaluation campaign of the 300W challenge.
Among these algorithms, Deng~\cite{deng2016m} and Fan~\cite{fan2016approaching} are the academia and industry winners of the challenge.

\begin{figure}[!t]
\centering
\subfloat[Semi-frontal]{
\label{fig_9_1}
 \includegraphics[width=.95\linewidth]{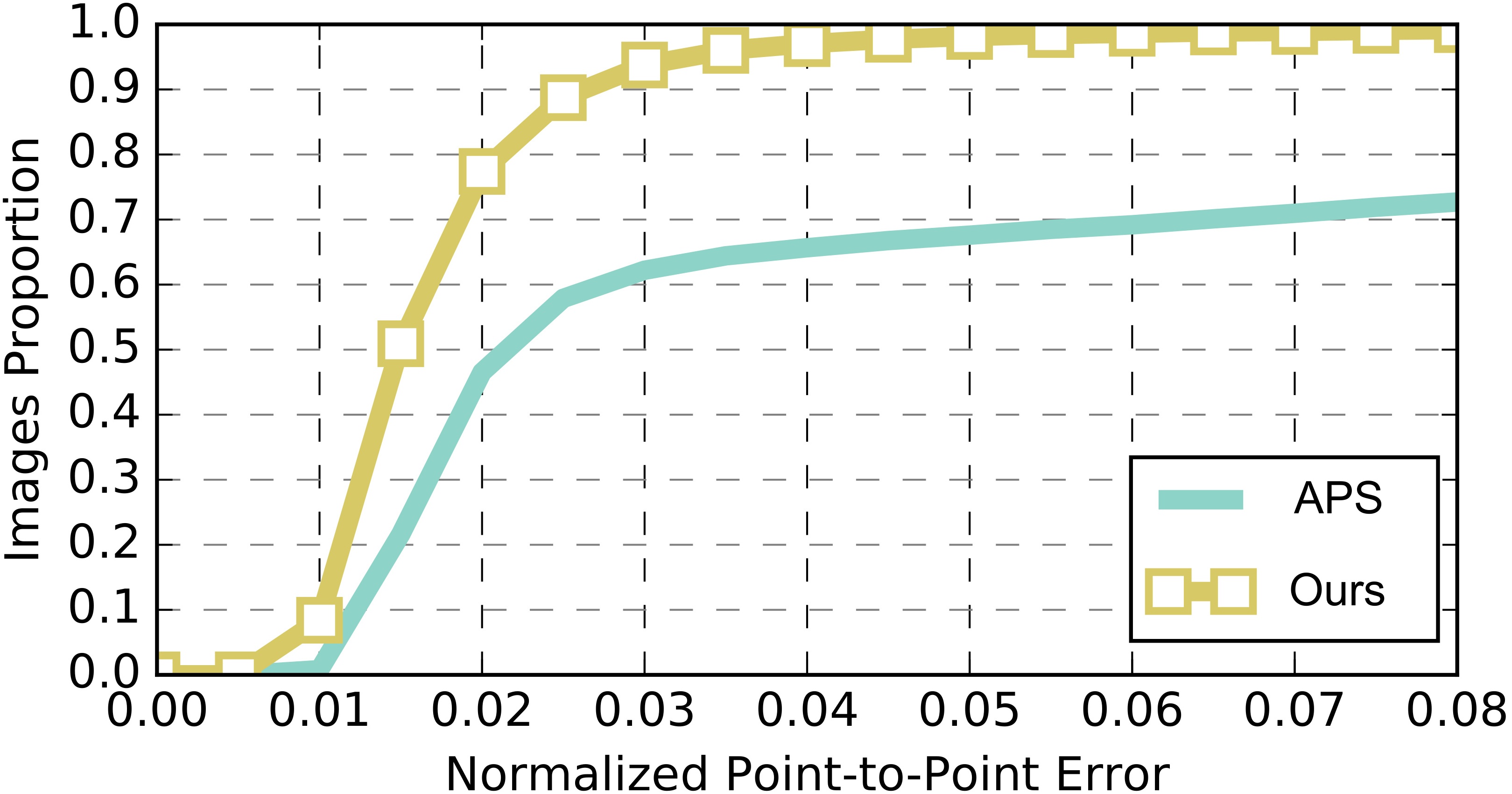}
}
\\
\subfloat[Profile]{
\label{fig_9_2}
 \includegraphics[width=.95\linewidth]{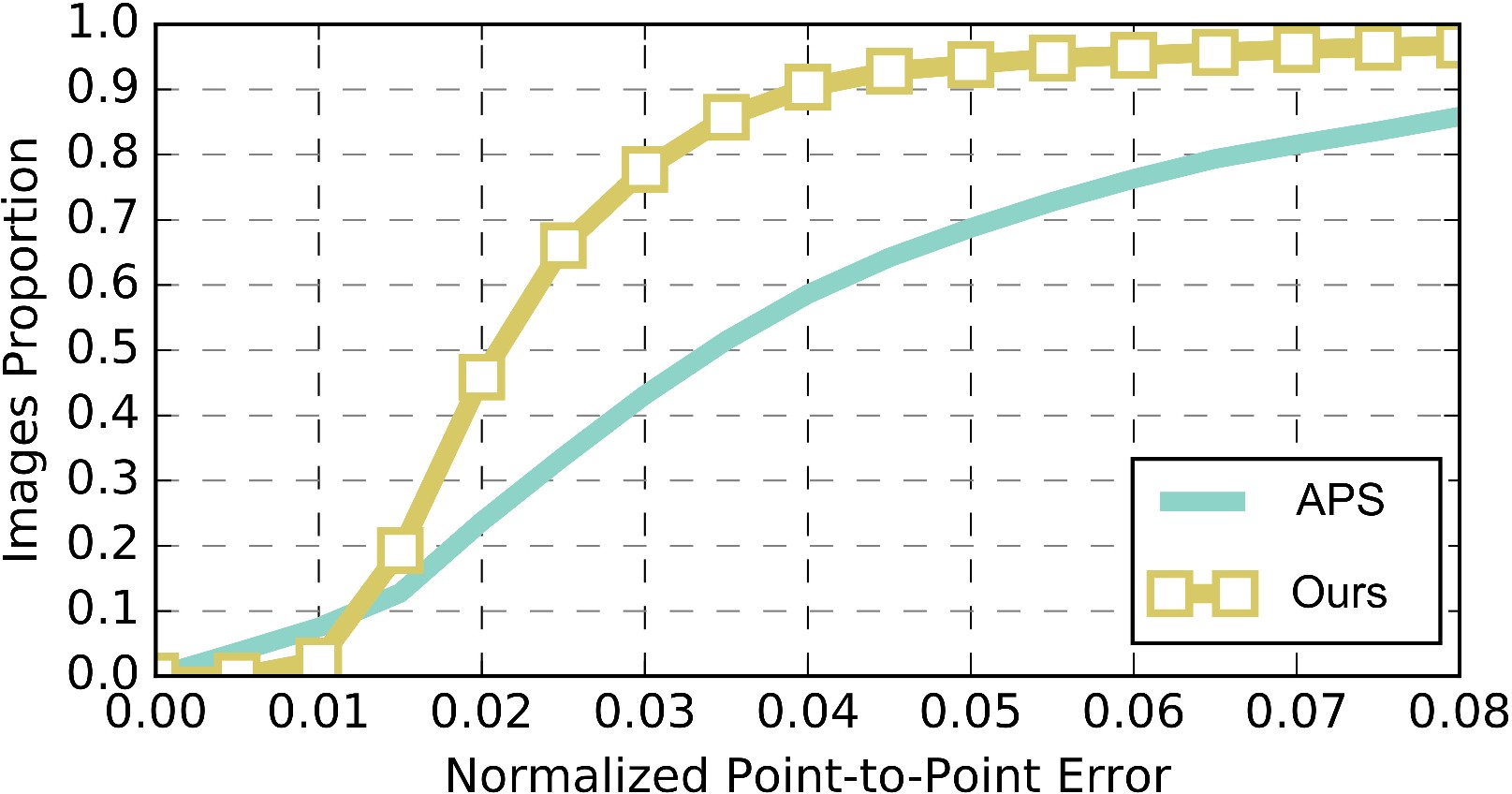}
}
\caption{A comparison between our proposed framework and the APS method on (a) the semi-frontal, and (b) the profile subsets of the Menpo benchmark.}
\label{fig_9}
\end{figure}

\begin{figure*}[th]
\centering
\subfloat[Semi-frontal]{
\label{fig_10_1}
\includegraphics[trim = 0mm 105mm 0mm 98mm, clip,width= 1\linewidth]{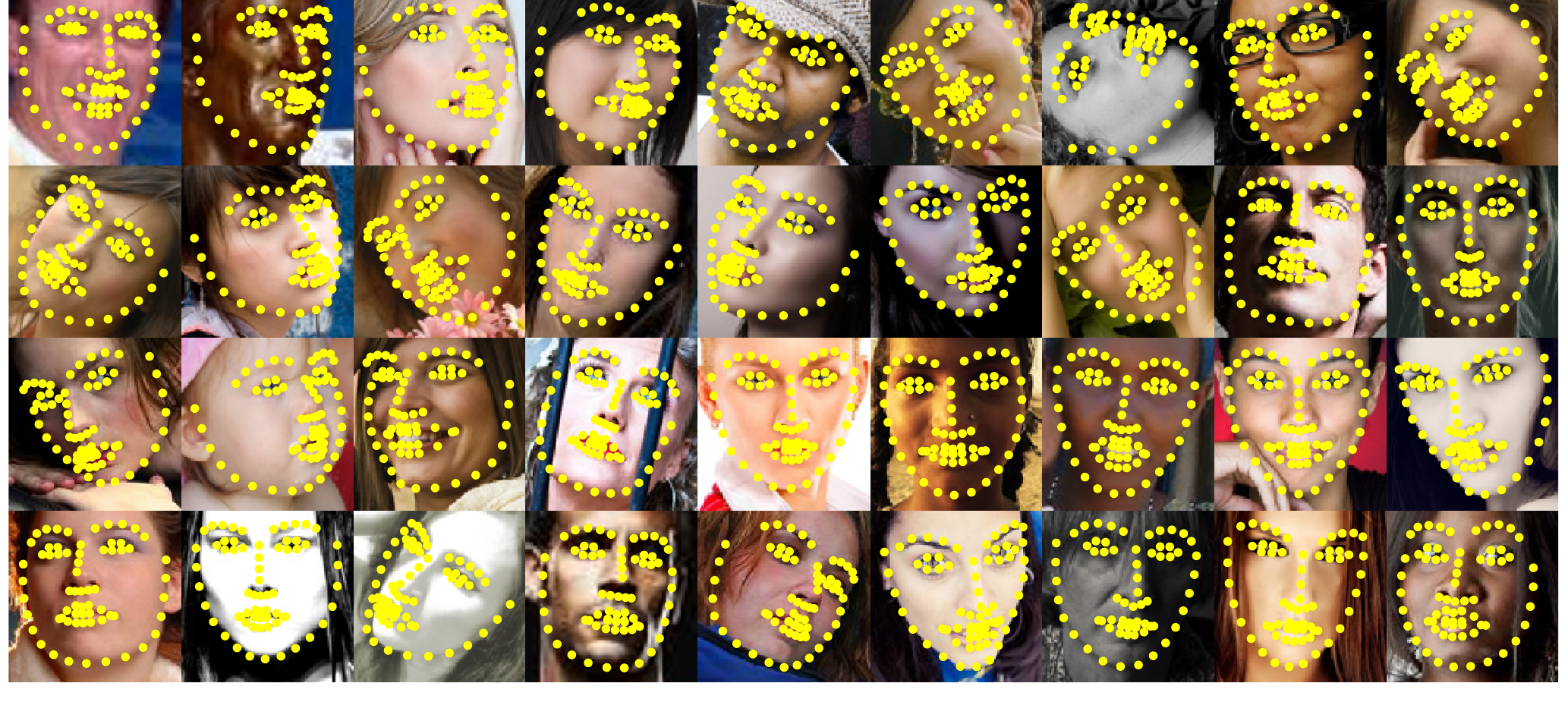}
}
\\
\subfloat[Profile]{
\label{fig_10_2}
\includegraphics[trim = 0mm 105mm 0mm 98mm, clip,width= 1\linewidth]{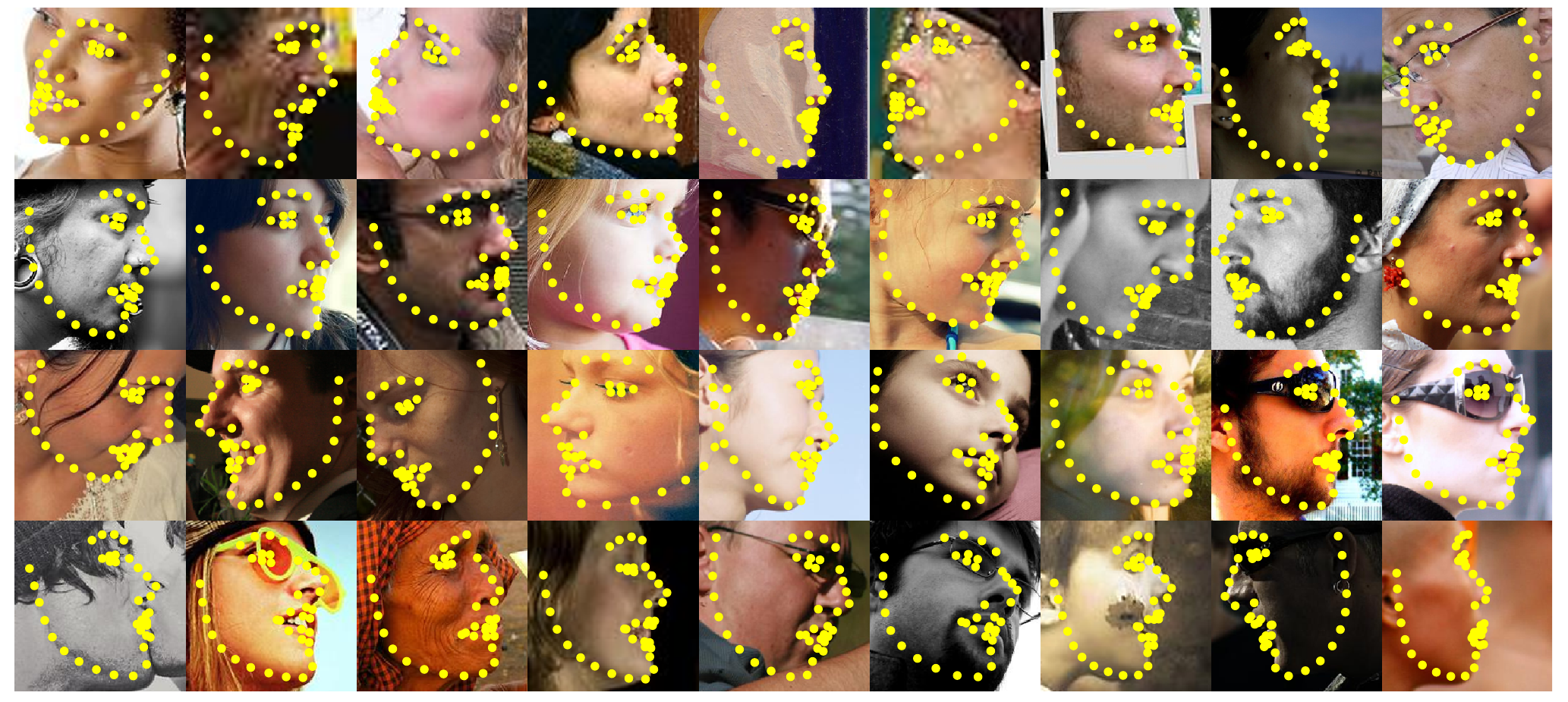}
}
\caption{Some landmark localisation results of the proposed method on the Menpo benchmark.}
\label{fig_10}
\end{figure*}
\subsection{Evaluation on the Menpo Benchmark}
The final facial landmark localisation results on the test set of the Menpo benchmark are shown in Fig.~\ref{fig_9}.
The figure compares our proposed framework with the baseline method, \ie Viola-Jones face detector and Active Pictorial Structures (APS)~\cite{antonakos2015active}.

The proposed framework significantly outperforms the baseline method in terms of accuracy.
It is clear that the face detection and bounding box aggregation techniques used in our approach successfully detected the majority of the test faces.
Moreover, the use of pose estimation and cascaded shape regression helps to achieve accurate facial landmark localisation results.
Some examples are shown in Fig.~\ref{fig_10}.

\section{Conclusion}
\label{sec_9}
We have presented a coarse-to-fine facial landmark localisation framework for the Menpo benchmark competition.
The proposed framework applied four face detectors and a bounding box aggregation method for robust and accurate face detection.
Then a cascaded shape regression model trained using a number of training samples with a wide spectrum of pose variations was used for rough facial landmark localisation and pose estimation.
Last, we rotated or flipped the test images for accurate facial landmark localisation using a fine-grained cascaded shape regression model trained using a face dataset with limited pose variations.

The experimental results carried out on the 300W and Menpo benchmarks demonstrate the superiority of the proposed framework.
The key to the success of the proposed framework is the splitting of the original cascaded shape regression process into a number of coarse-to-fine steps.
In addition, the use of an ensemble of multiple face detectors greatly improves the accuracy of the face detection step.

\section*{Acknowledgements}
This work was supported in part by the EPSRC Programme Grant `FACER2VM' (EP/N007743/1), the National Natural Science Foundation of China (61373055, 61672265) and the Natural Science Foundation of Jiangsu Province (BK20140419, BK20161135).

{\small
\bibliographystyle{ieee}
\bibliography{mybib}
}

\end{document}